\documentclass[conference]{IEEEtran}

\usepackage{lineno}

\usepackage{microtype}
\usepackage{graphicx}
\usepackage{colortbl}
\usepackage{booktabs} 
\usepackage{tabularx}
\usepackage[all]{nowidow}

\usepackage[dvipsnames]{xcolor}

\usepackage[draft]{hyperref}

\usepackage{todonotes}

\usepackage{algorithm, algorithmic}

\usepackage{epsfig}
\usepackage{amsmath}
\usepackage{amssymb}
\usepackage{subfiles}
\usepackage{commath}
\usepackage{multirow}

\newcommand{\argmin}{\mathop{\mathrm{argmin}}\limits}
\usepackage{subfigure}

\usepackage{afterpage}

\begin{document}

\title{Composition of Saliency Metrics for Pruning with a Myopic Oracle}

\author{
\IEEEauthorblockN{Kaveena Persand}
\IEEEauthorblockA{\textit{School of Computer Science} \\
\textit{Trinity College Dublin}\\
Dublin 2, Ireland \\
persandk@tcd.ie}
\and
\IEEEauthorblockN{Andrew Anderson}
\IEEEauthorblockA{\textit{School of Computer Science} \\
\textit{Trinity College Dublin}\\
Dublin 2, Ireland \\
aanderso@tcd.ie}
\and
\IEEEauthorblockN{David Gregg}
\IEEEauthorblockA{\textit{School of Computer Science} \\
\textit{Trinity College Dublin}\\
Dublin 2, Ireland \\
david.gregg@cs.tcd.ie}
}

\maketitle
\IEEEpubidadjcol

\begin{abstract}

The cost of Convolutional Neural Network (CNN) inference can be reduced by
\emph{pruning} weights from a trained network, eliminating computations while
preserving the predictive accuracy up to some threshold. While many heuristic
\emph{saliency metrics} have been proposed to guide this process, the quality of
pruning decisions made by any one metric is highly context-sensitive. Some
metrics make excellent pruning decisions for one network, while making poor
decisions for other networks.

Traditionally, a single heuristic saliency metric is used for the entire pruning
process. We show how to \emph{compose} a set of these saliency metrics to form a
much more robust (albeit still heuristic) saliency. The key idea is to
exploit the cases where the different base metrics do well, and avoid the cases
where they do poorly by switching to a different metric. With an experimental
evaluation of channel pruning on several popular CNNs on the CIFAR-10 and
CIFAR-100 datasets, we show that the composite saliency metrics derived by our
method consistently outperform all of the individual constituent metrics.

\end{abstract}


\section{Motivation}

The fundamental action of \emph{pruning} is to identify a subset of the weights
of a neural network which can be removed (pruned) without damaging the
predictive accuracy of the network by more than a user-supplied threshold. This
threshold is application specific: in optical character recognition, for
example, the acceptable drop in accuracy may be much larger than in real-time
pedestrian detection.

In order to find optimal prunings of a neural network, we can use a simple, but
computationally expensive approach: considering all possible subsets of
unpruned weights, prune each subset independently and measure the change in
accuracy to select the optimal set of weights to remove.  In practice, pruning
is not done in this way due to the prohibitive cost.  Instead a pruning scheme
which determines when, how many, and which weights need to be removed is used.

Many pruning schemes use a saliency metric to determine which subsets of
weights are least likely to damage the network\cite{DSD,LWC, Dally, Hu, Lecun,
Hassibi, Mozer_Smolensky}.  The saliency metric is a heuristic used
to efficiently rank different subsets of weights. A wide variety of heuristic
saliency metrics have been proposed over decades of research in artificial
intelligence\cite{LWC, Lecun, Hassibi, Mozer_Smolensky, Molchanov}. Each of
these heuristic saliency metrics may perform better or worse in context, with
no one metric being clearly superior. When pruning a neural network, the data
scientist must often resort to simple rules of thumb or guesswork to select an
appropriate heuristic to guide the pruning process.

\subsection{Contributions}

We propose a method to derive a \emph{composite saliency metric} which can avoid
poor choices made by otherwise effective \emph{constituent} saliency metrics.
Our approach uses a \textit{myopic oracle} to decide which of a fixed set of
constituent metrics should be active at every step of the pruning process. As
the predictive power of the constituent metrics waxes and wanes, our approach
dynamically switches between metrics so that the most appropriate metric is
guiding the process at all points.

We make the following principal contributions:
\begin{itemize}
\item We demonstrate that different metrics perform differently on different networks.
\item We show how to compose different saliency metrics automatically (the myopic oracle).
\item We experimentally investigate fusion of state-of-the-art saliency metrics.
\end{itemize}

\section{Background} \label{sec:background}

The goal of pruning is to remove the largest structured or unstructured group
of weights without damaging the performance of the network.  Pruning schemes
can be categorised roughly as either saliency based or penalty term based.
Penalty term based schemes modify the cost function to prune weights. Saliency
metric based schemes use a metric to identify which weights or set of weights
are least important to the network.  While early work on saliency metrics
focused on using fully trained networks~\cite{Mozer_Smolensky,Karnin90,Lecun,
Hassibi}, recent work has shown that saliency metrics can successfully be used
to remove parameters from the network at different stages of the training
process~\cite{Frankle_Carbin, SNIP}.

In this paper, we focus on saliency metrics for coarse granularity pruning,
i.e., removing entire convolution channels from a network.

\subsection{Classification of Metrics}

In the following treatment, we refer to the $c$th subset of
weights as $W_c$, with corresponding output activations $A_c$. Weight subsets
may have an arbitrary positive nonzero size ${\lVert W_c \rVert}_0 $.

In order to obtain the activation values $A_c$, a forward pass of the
network is required, since activations are dynamic information. Weights are
static information, and are immediately available without performing any
computation. This distinction allows us to neatly categorize many pruning
metrics which have been proposed in the literature.

\begin{table}[h]
\caption{Notable saliency metrics used for channel pruning.}
\label{tab:saliency_equations}
\begin{center}
\begin{tabular}{m{3.5cm}l}
\textbf{Saliency Metric} & \textbf{Equation} \\
\midrule
Mean squares of weights \cite{Molchanov} &
\parbox{4cm}{ \begin{equation}\label{eqn:mean_sqr_weights} S_c = \frac{1}{\lVert{W_c \rVert}_0} \sum_{w \in W_c} w^2 \end{equation}}
\\
Mean of activations \cite{Anwar} &
\parbox{4cm}{ \begin{equation}\label{eqn:mean_abs_act} S_c = \frac{1}{{\lVert A_c \rVert}_0}\sum_{a \in A_c}  a \end{equation}}
\\
Average of gradients \cite{Liu2019} &
\parbox{4cm}{ \begin{equation}\label{eqn:mean_gradients} S_c = \frac{1}{{\lVert A_c \rVert}_0 } \left \lvert\sum_{a \in A_c} \frac{d\mathcal{L}}{da} \right \lvert \end{equation}}
\\
1\textsuperscript{st} order Taylor expansion \cite{Molchanov} &
\parbox{4cm}{ \begin{equation}\label{eqn:taylor} S_c = \frac{1}{{\lVert A_ \rVert}_0} \left \lvert\sum_{a \in A_c} a \frac{d\mathcal{L}}{da} \right \lvert \end{equation}}
\\
2\textsuperscript{nd} order Taylor expansion using Fisher information \cite{Theis} &
\parbox{4cm}{ \begin{equation}\label{eqn:fisher} S_c =  \frac{1}{2} \left ( \sum_{a \in A_c} a \frac{d\mathcal{L}}{da} \right ) ^{2} \end{equation}}
\\
\end{tabular}
\end{center}
\end{table}

\subsubsection{Weight-based Saliency Metrics}

Commonly used \textit{static} saliency metrics are the L1-norm
of the weights \cite{Dally} and the mean squares of the weights\cite{Molchanov} (see
Equation \ref{eqn:mean_sqr_weights}).
The L2 and L1 norm of the weights have been used in multiple pruning schemes
\cite{GroupWiseBrainDamage,Li2017,SoftFilterPruning} for different
granularities of pruning.  Weights-based saliency metrics assume that weights
of lower magnitude have a lower contribution to the network.

\subsubsection{Activation-based Saliency Metrics}

More recent work has proposed \textit{dynamic} saliency metrics, which can exploit
the information in the activations and gradients. These can only be obtained by
performing a forward, and an additional backward, pass of the network,
respectively.

Some examples of effective dynamic saliency metrics are the
absolute-percentage-of-zeros \cite{Hu}, mean \cite{Anwar} (Equation
\ref{eqn:mean_abs_act}) and standard deviation of activations
\cite{Polyak_Wolf}.  Activation-based saliency metrics exploit information only
obtainable during forward passes of the network.

\subsubsection{Gradient-based Saliency Metrics}
Conversely we can find saliency metrics that make use of only the
gradients \cite{Karnin90} such as the use of the average of the gradients, \cite{Liu2019}
(Equation \ref{eqn:mean_gradients}).

However, the information contained in the gradients is often coupled with the
activations. The 2\textsuperscript{nd} order Taylor expansion using Fisher
information \cite{Theis} (Equation \ref{eqn:fisher}) and 1\textsuperscript{st}
order Taylor expansion \cite{Molchanov} (Equation \ref{eqn:taylor}) are two
notable examples of saliency metrics combining both the information of the
activations and their gradients.

\subsection{Metric Assumptions}

\begin{figure}[h]
\centering
\includegraphics[width=0.95\columnwidth]{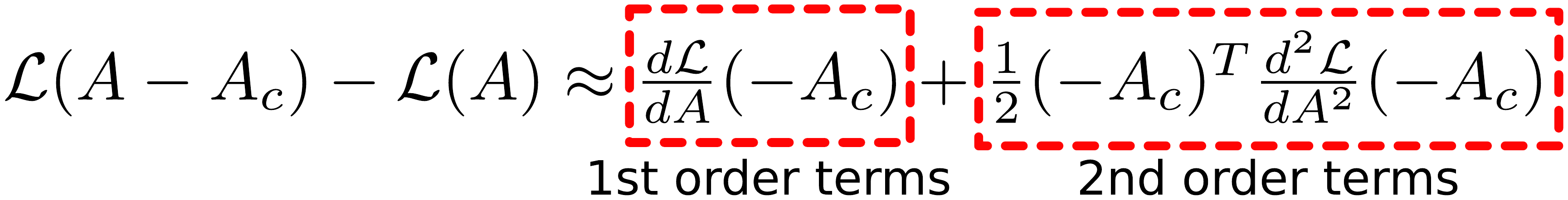}
\caption{Estimating the effect of pruning the $c^{th}$ output channel of a
network with loss function $\mathcal{L}$ using a 2\textsuperscript{nd} order Taylor development
around $A$, the activations of the network.}
\label{fig:taylor_expansion}
\end{figure}

Most of these saliency metrics rely on some assumptions.  For example, when
using the L1-norm of weights, the assumption is that smaller weights contribute
less to the network.  Their underlying assumptions can sometimes be
conflicting.  The 2\textsuperscript{nd} order Taylor expansion using Fisher
information and 1\textsuperscript{st} order Taylor expansion are both derived
using the Taylor expansion presented in Figure \ref{fig:taylor_expansion}.
However, they are constructed under different assumptions. The construction of
the 2\textsuperscript{nd} order Taylor Expansion Using Fisher Information
\cite{Theis} assumes that the gradients of the weights and activations are
insignificant.  Hence, the $1^{st}$ order terms in Figure
\ref{fig:taylor_expansion} are ignored and the second order terms are
approximated to the Fisher information to derive Equation \ref{eqn:fisher}.  On
the other hand, when using a first order Taylor expansion, the higher order
terms are considered insignificant, meaning we ignore the $2^{nd}$ order terms
in Figure \ref{fig:taylor_expansion} to derive Equation \ref{eqn:taylor}.

This is a crucial distinction, because these built-in assumptions in the
construction of the metrics are typically \emph{not simultaneously true} for
any given network.  Moreover, as the pruning process continues, the
\emph{degree} of significance of different components can, and does, change.
When most of the remaining information is in higher order components, metrics
using only first order components are effectively making random decisions, and
vice versa.

For example, in the case of pruning a partially converged network, the
gradients of the weights and activations are very unlikely to be negligible.
When pruning a fully converged network, the gradients are much more likely to
be negligible. When pruning a fully trained network, we start with converged
weights. However, as the pruning process proceeds, we may end up pruning
\emph{partially} converged weights, since the pruning process degrades the
network.

\section{Composing Saliency Metrics} \label{sec:composing_saliency_metrics}

\begin{figure*}[t]
\centering
\subfigure
[Channel selection ($k = 3$) and oracle ~~~~~~~ evaluation.]{\label{fig:myopic_oracle2}\includegraphics[width=0.39\linewidth]{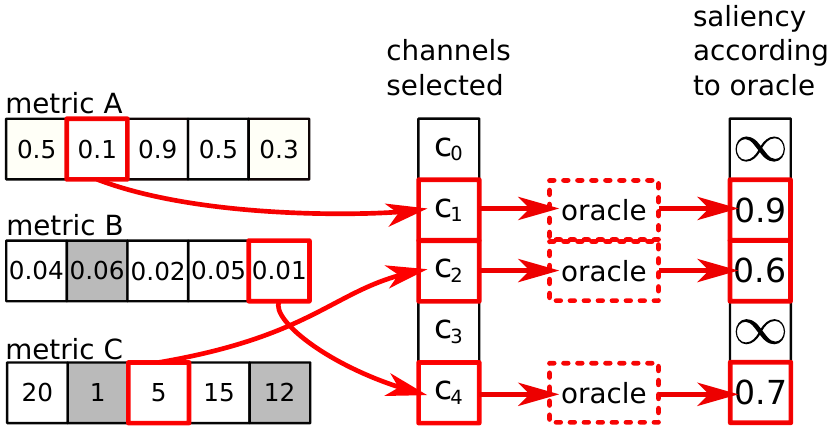}}
\subfigure
[Sensitivity evaluation by the oracle for a channel, $c$, according to Equation \ref{eqn:sensitivity}.]{\label{fig:myopic_oracle3}\includegraphics[width=0.6\linewidth]{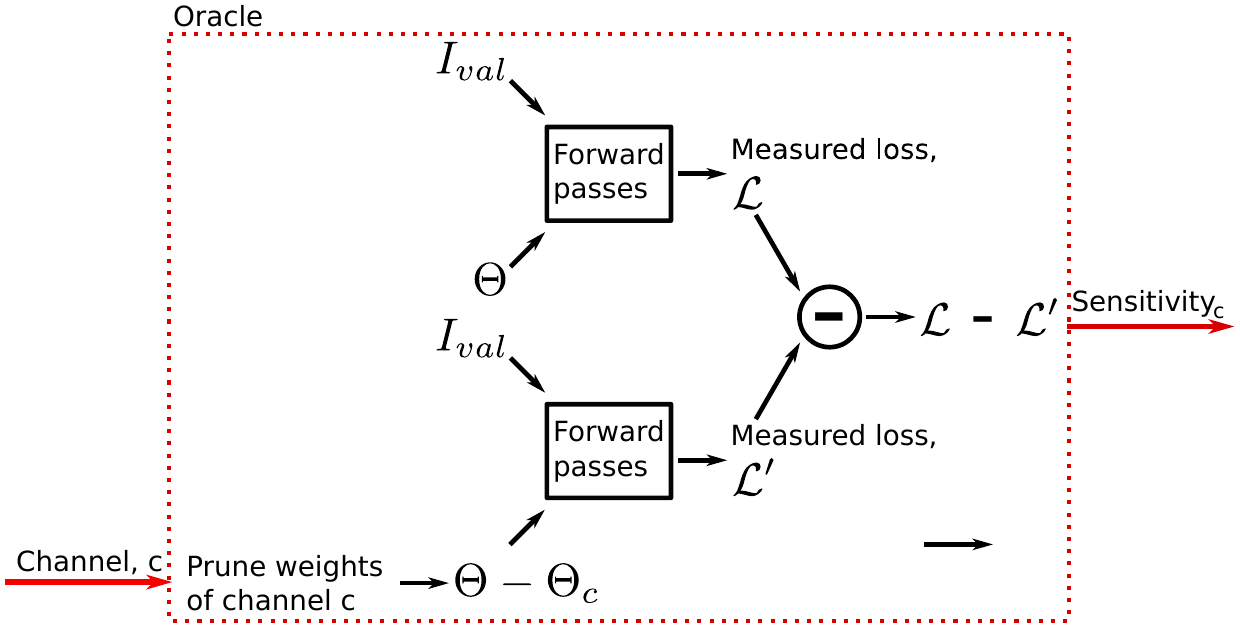}}
\caption{Combining rankings of saliency metrics A, B and C using a myopic oracle with $k=3$.}
\label{fig:myopic_oracle}
\end{figure*}

When using any one saliency metric for the entire pruning process, we run the
risk of the metric assumptions being invalidated, leading to poor decisions
being made by the metric.  Ideally we could combine the best aspects of
different saliency metrics.  The chief difficulty lies in combining the
numerical output of different pruning metrics, which are not directly
comparable.

Consider the application of two saliency metrics $A$ and $B$ to neural network
$\mathcal{N}$ with a total of four weights $[a, b, c, d]$. Let us suppose that
ranking the weights with each metric yields the rankings $A(\mathcal{N}) =
[0.5, 0.4, 0.9, 0.1]$ and $B(\mathcal{N}) = [0.01, 0.05, 0.04, 0.06]$. Metric
$A$ indicates removing weight $d$ will have the least effect, while metric $B$
indicates removing weight $a$ will have least effect.

If the metrics agree on the weights to be removed, there is no issue. However,
if they disagree, there are two alternatives to consider. If either choice
results in the same amount of damage to the network we can call the
disagreement trivial. However, if one causes more damage than the other, the
disagreement is nontrivial. If we continue to use the suboptimal metric to
guide the process, we will introduce more and more relative error.

Inspection of the numerical \emph{saliency} values assigned to each weight or
set of weights by metrics $A$ and $B$ exposes the difficulty of combining these
metrics numerically. Although each metric ranks the weights or set of weights
in the network, in principle, these saliency values can have arbitrary scales.
If we were to combine the metrics with a simple linear combination, such as a
weighted average, one metric would be disproportionately selected. As we add
more and more metrics to the set, the difficulty increases. However, by
performing a forward pass of the network, we may determine at any point the
true effect on the loss function of making a particular pruning decision. This
is the key to our proposed approach.

\subsection{Our Proposed Method: Myopic Oracle} \label{sec:proposed_method}

When different saliency metrics yield different rankings for the same sets of
weights, we can evaluate which ranking is the most correct by performing a
direct measurement of the \emph{sensitivity} of the network to the removal of
the proposed subsets of parameters.

For brevity of presentation, in the following treatment we chose subsets of
parameters corresponding to whole output feature maps (i.e.  \emph{channel
pruning}). However, the approach is no different for other subsets of weights
(filters, individual weights or any other granularity of pruning).

For a CNN characterised by the loss function $\mathcal{L}$ and permanent
weights $\Theta$, the sensitivity of the $c^{th}$ channel of the network, using
forward passes on the validation set $I_{val}$, is given by the change in the
loss caused by replacing all the weights of the $c^{th}$ channel $\Theta_c$
with zeros as given in Equation~\ref{eqn:sensitivity}.

In the case of channel pruning, $\Theta_c$ denotes all the parameters that need
to be removed to remove a channel and still end up with a dense network.
Hence, $\Theta_c$ contains $W_c$ but may include parameters from other layers
that interact with $c^{th}$ channel.
\begin{equation}
\label{eqn:sensitivity}
Sensitivity_c = \mathcal{L}(\Theta - \Theta_c, I_{val}) - \mathcal{L}(\Theta, I_{val})
\end{equation}

At every pruning step, the \textbf{myopic oracle} measures the sensitivity of
only $k$ different channels using the validation set. Notionally, $k$ is the
number of channels that the myopic oracle can ``see". The choice of the value
of $k$ depends on the pruning scheme used, but must be at least the number of
channels that the pruning scheme considers pruning simultaneously. Hence, $k$
can vary depending on the pruning scheme.

It should be noted that the different consituent saliencies and the sensitivity
computed by the myopic oracle use the same dataset, $I_{val}$, containing
$N_{val}$ batches of images.
The cost of running the myopic oracle for one channel is similar to the cost of
computing the dynamic heuristics that use forward passes only.  Hence, if
$N_{val}$ batches are used to measure the sensitivity for each channel the cost
of running the oracle (excluding the cost of computing the individual
saliencies) is $ k \times N_{val} \times cost~of~forward~pass$. Assuming that
the cost of a forward pass is roughly equal to the cost of a backward pass, the
cost of computing a gradients-based saliency metric also using $N_{val}$
batches is $2 \times N_{val} \times cost~of~forward~pass$.  The cost of the
myopic oracle is hence not prohibitive but needs to be factored when choosing
$k$. A wider view may yield better results but at an increased computational
cost.

The myopic oracle visits each of the constituent saliency metrics in a
round-robin fashion, and selects the \textbf{lowest ranked channel} to add to
the set of channels whose sensitivity should be measured. If the lowest ranked
channel has already been selected by another constituent, the second lowest is
used instead, and so on. This process continues until $k$ unique channels have
been selected. The sensitivity of each channel is then tested, yielding the
true ranking of these $k$ channels.

Note that the actual saliency values output by each saliency metric are never
consumed by the oracle: only the implied ordering of the channels is used. In
this way, the oracle is agnostic to the scales of the individual pruning
metrics.

Figure~\ref{fig:myopic_oracle2} illustrates selection of channels to be
evaluated by the oracle in the case of a pruning scheme with $k=3$.  The
selected channels then have their sensitivities measured by the oracle
according to Equation \ref{eqn:sensitivity} and Figure \ref{fig:myopic_oracle3}.

\subsection{Constituent Saliency Metrics}

Our composite approach can be used with any saliency metric which can be
expressed as a function of weights and activations (including all gradients,
which are derivatives of one with respect to the other). However, composing all
published saliency metrics following this schema would be unrealistic. Instead
we choose a sample of prominent saliency metrics from the literature that
perform well in practice. These metrics rely on different kinds of information.
We consider both static and dynamic saliency metrics.

We selected the constituent saliency metrics shown in Table
\ref{tab:saliency_equations} to be combined via the myopic oracle. Prior work
has shown each of these saliency metrics are very effective.

Even though they are known to perform well, the chosen saliency metrics are
constructed under different assumptions and use a diverse selection of
parameters from the network.

\section{Experimental Setup} \label{sec:setup}

For our experimental setup, we chose a set of constituent saliency metrics in
Table \ref{tab:saliency_equations} to compose via the myopic oracle, and also a
general \emph{pruning scheme} to follow. Simple pruning schemes rely heavily on
the saliency metric's prediction whereas in sophisticated schemes\cite{AMC,
DingDGHY19, GateDecorator, ADMM_Regulizer}, the contribution of the saliency
metric can become obfuscated by other factors.

\subsection{Choice of Pruning Scheme}

Since our objective is specifically to study the \textbf{differences in pruning
metrics}, we chose to eliminate confounding factors by using a simple,
iterative pruning scheme \textbf{without fine-tuning or retraining}. The only
change we make to the network weights is to set pruned weights to zero.  Using
a pruning scheme with retraining is needed to find the absolute best network,
however introducing retraining introduces more stochasticity in the results.
Since our aim is not to find the best network but the best saliency metric,
retraining can obsfucate the results.

Even when retraining is in use, saliency metrics which cause less deviation
from the initial test accuracy can lead to less time being spent on retraining,
and also to large groups of channels being simultaneously removed, in the case
of pruning schemes that allow for simultaneous pruning of multiple channels.
Hence, a better saliency metric will always reduce the total amount of effort
used to produce pruned networks. Algorithm \ref{alg:experiment} outlines the
simple pruning scheme used to evaluate the myopic oracle.

\begin{algorithm}[h]
   \caption{Evaluating different channel selections for a CNN with loss
   function $\mathcal{L}$, accuracy $\mathcal{Y}$ and converged weights
   $\Theta$ with $M$ channels for a user-defined maximum drop in initial test accuracy, $maxTestAccDrop$}
   \label{alg:experiment}
\begin{algorithmic}
  \STATE $initialTestAcc \gets \mathcal{Y}(\Theta, I_{test})$
  \REPEAT
    \STATE $S_c \gets computeSaliency(\mathcal{L}, \Theta, \Theta_c, I_{val})$ for $ c \in \{ 0 \ldots M-1 \}$
    \STATE Get $j$, such that $j = \argmin_{c \in \{0 \ldots M-1\}} S_c$  and $\Theta_j$ is a non-zero vector.
    \STATE $\Theta \gets \Theta - \Theta_j$
    \STATE $testAcc \gets \mathcal{Y}(\Theta, I_{test})$
  \UNTIL{$testAcc < initialTestAcc - maxTestAccDrop$}
\end{algorithmic}
\end{algorithm}

We iteratively recompute the channel which should be removed, and remove one
channel at a time from the entire network until the test accuracy is degraded
beyond a certain threshold.

\subsubsection{CNN Models}

LeNet-5~\cite{lenet-paper} and AlexNet~\cite{alexnet} are modified so that the
first convolutions accept $32 \times 32$ RGB input images and classify the
images between 10 categories.  ResNet-20~\cite{resnet-paper},
NIN~\cite{nin-paper} and the CIFAR10~\cite{cifar10} network are used according
to their original descriptions for the CIFAR10 dataset.  We also adapted
ResNet-20, NIN and AlexNet for the CIFAR-100~\cite{cifar10} dataset. The
networks used are trained from scratch using Caffe~\cite{caffe}.

\begin{table}[h]
\caption{Summary of trained network accuracy on CIFAR-10 and CIFAR-100.}
\label{tab:sum_test_acc_cifar10}
\begin{center}
\begin{tabular}{m{1.42cm}|m{0.97cm}m{0.95cm}m{1.24cm}m{0.6cm}m{0.7cm}}
\textbf{Network} & \textbf{LeNet-5 } & \textbf{CIFAR10 } & \textbf{ResNet-20 } & \textbf{NIN } & \textbf{AlexNet } \\
\midrule
\textbf{Accuracy (CIFAR-10)}        & 69\%    & 73\%    & 88\%    & 88\%    & 84\%   \\
\textbf{Accuracy \mbox{(CIFAR-100)}}        & -    & -    &  59.2\%  & 65.7\%  & 54.2\%  \\
\textbf{Convolution weights}  & 26.5K   & 79.2.K  & 270K    & 966K    & 2.3M   \\
\end{tabular}
\end{center}
\end{table}

\begin{figure*}[h]
\centering
\subfigure
[ResNet-20]{\label{fig:resnet_20}\includegraphics[width=0.47\linewidth]{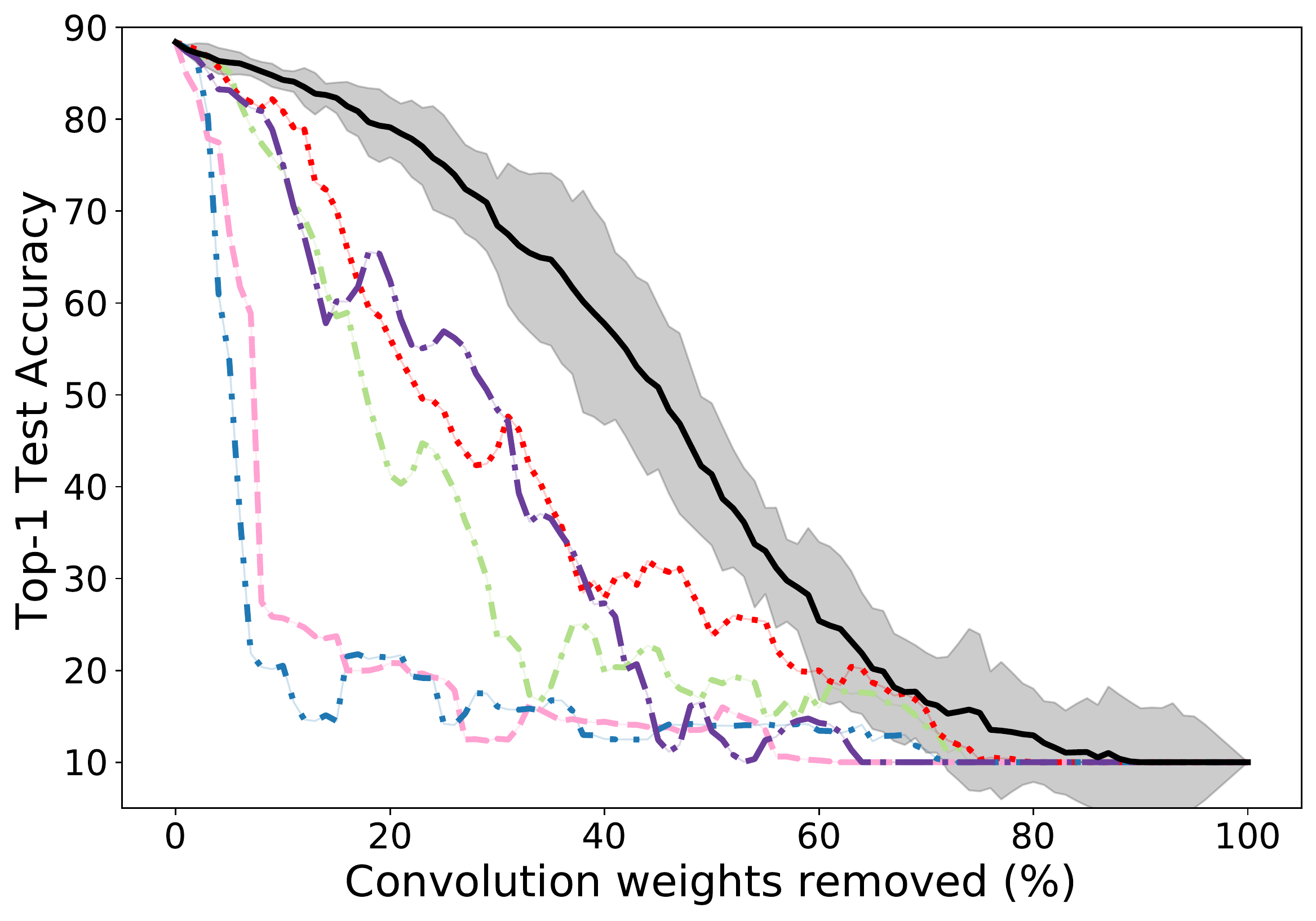}}
\subfigure
[NIN]{\label{fig:nin}\includegraphics[width=0.47\linewidth]{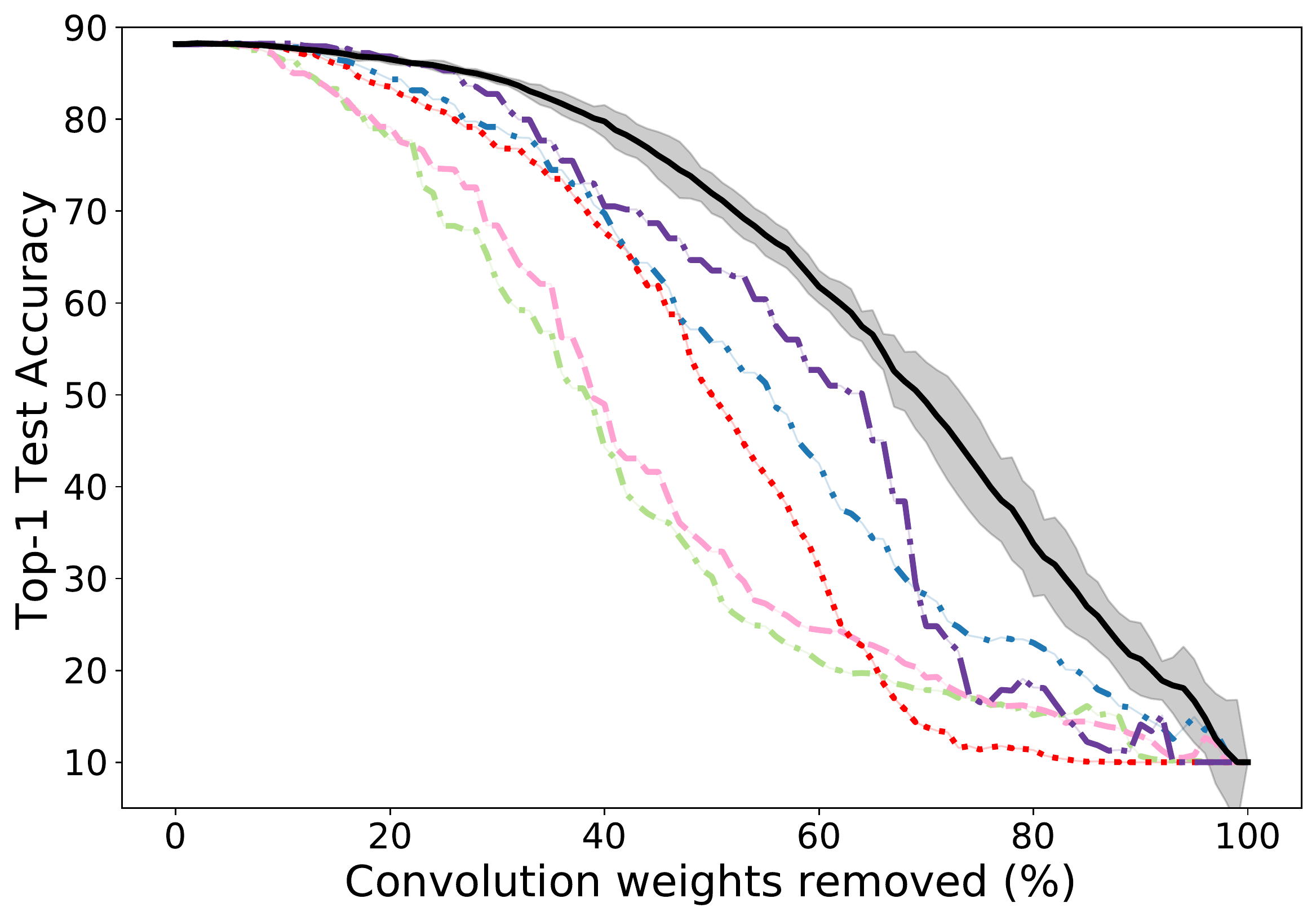}}
\subfigure
[LeNet-5]{\label{fig:lenet_5}\includegraphics[width=0.32\linewidth]{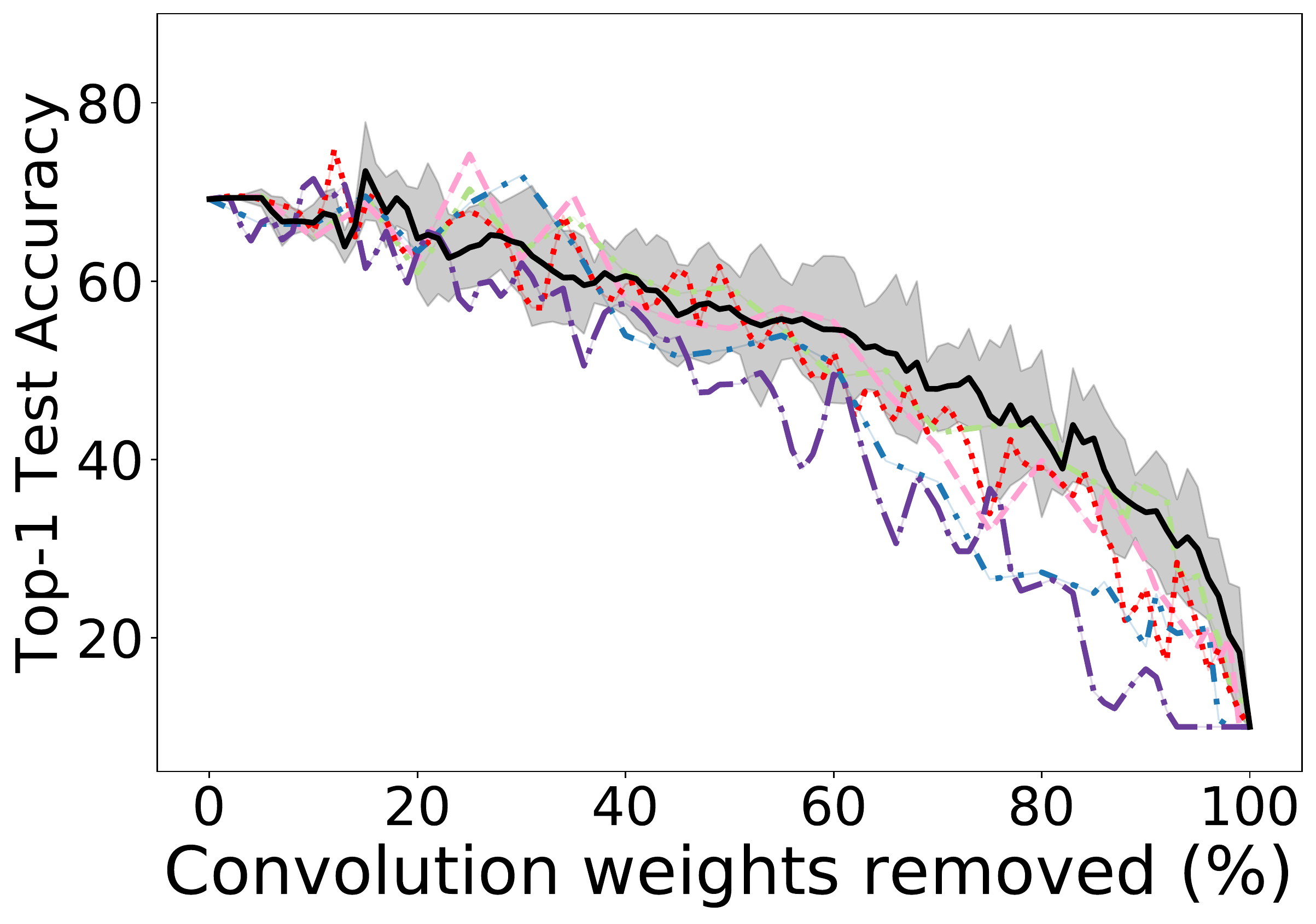}}
\subfigure
[CIFAR10]{\label{fig:cifar10}\includegraphics[width=0.32\linewidth]{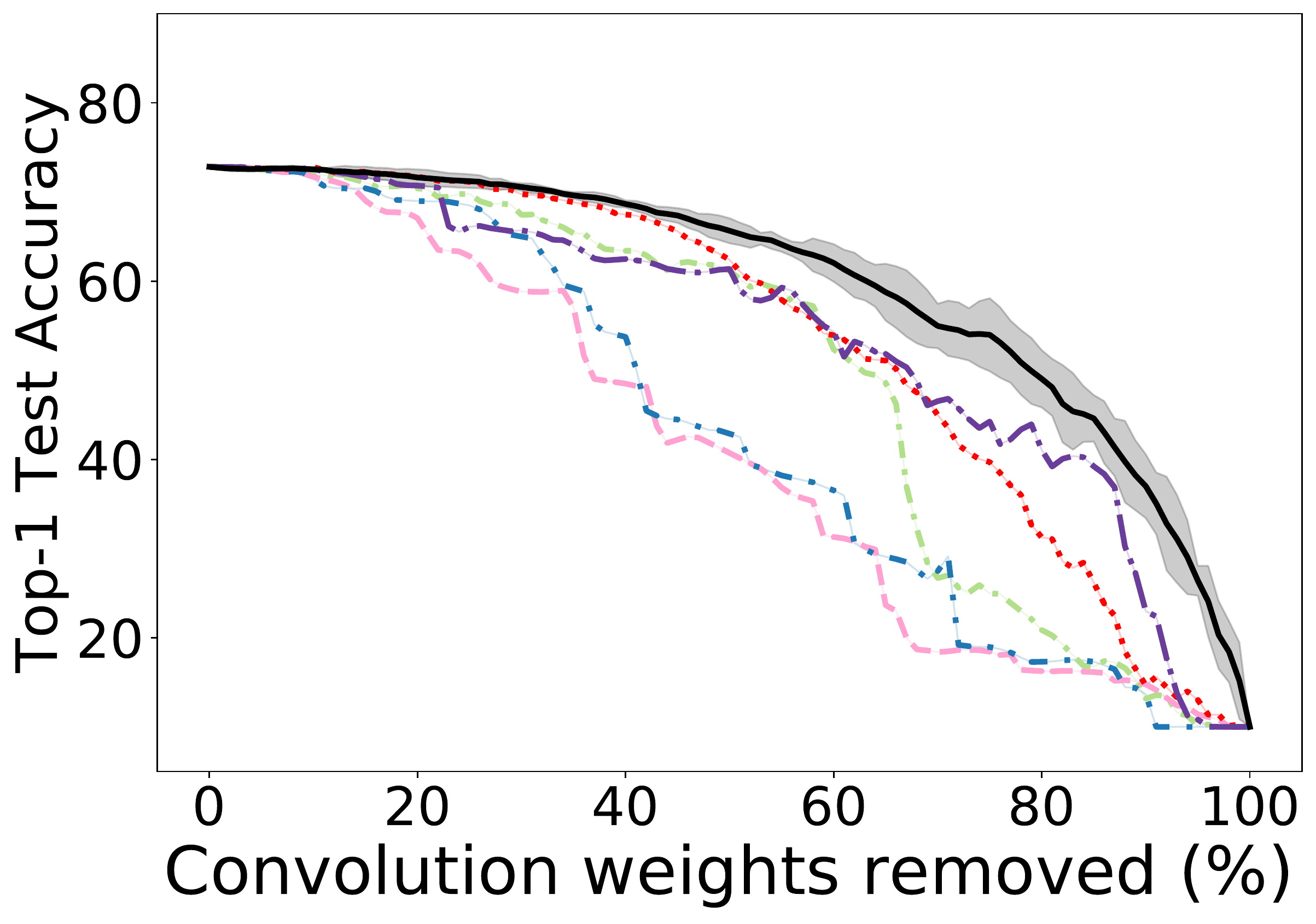}}
\subfigure
[AlexNet]{\label{fig:alexnet}\includegraphics[width=0.32\linewidth]{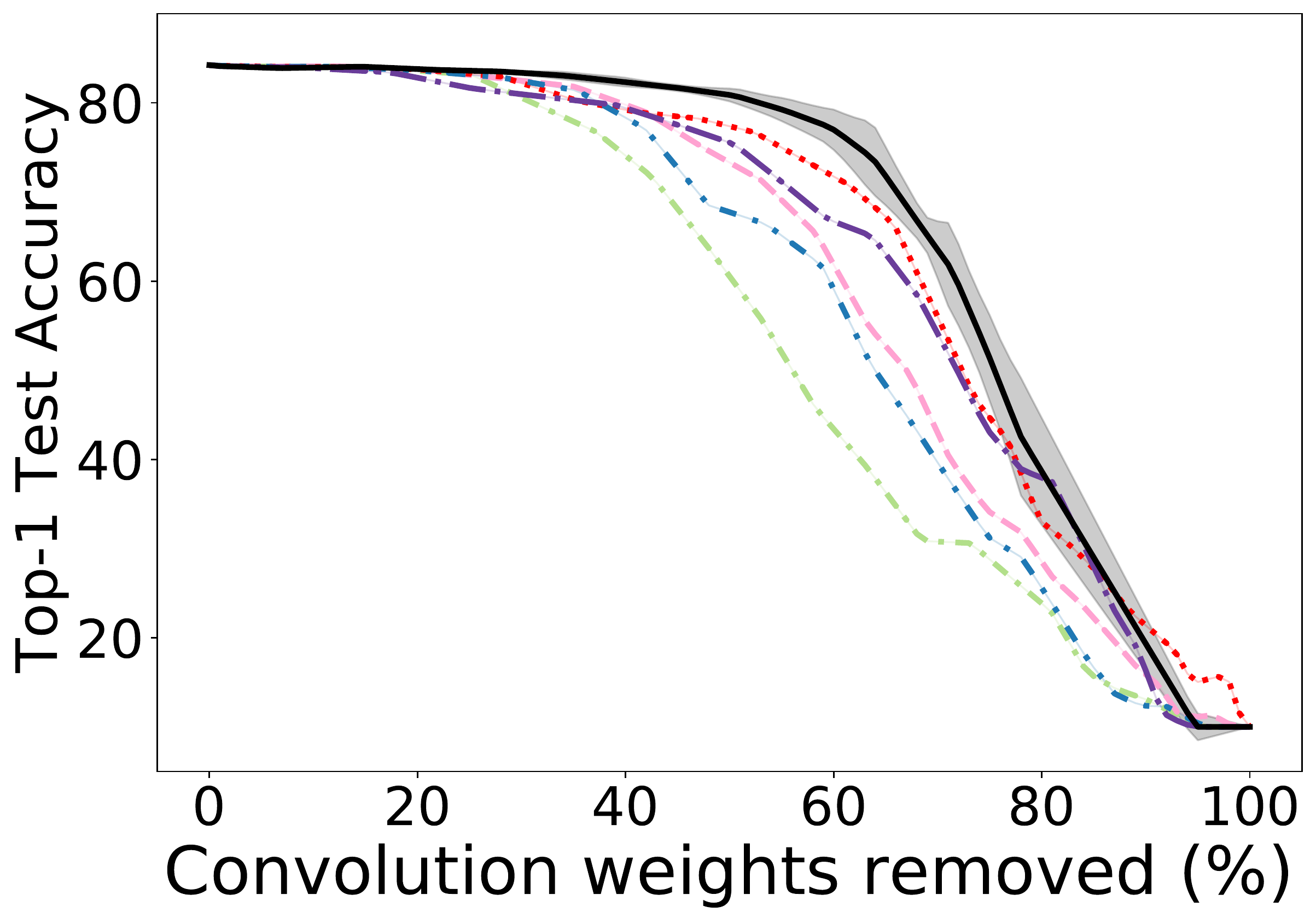}}
\subfigure
{\label{fig:legend}\includegraphics[width=0.9999999999\linewidth]{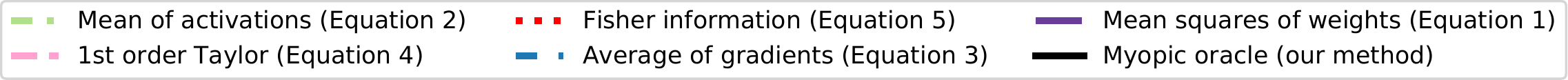}}
\caption{Graphs show top-1 test accuracy versus number of convolution
weights (\%) removed by pruning using the CIFAR-10 dataset. Individual saliency metrics are indicated
with dashed lines, and the myopic oracle (with $k=8$) is indicated with a solid line. Error
bands for the myopic oracle are shown based on a 95\%
confidence interval for 8 runs of the experiment.}
\label{fig:sensitivity_results}
\end{figure*}

\subsubsection{Experimental Setup and Hyperparameters}

We use the CIFAR-10 and CIFAR-100 datasets in our experiments using their
respective 50K/10K train/test images. We first train all the networks to
convergence using the full training set.  The accuracy of our trained networks
are shown in Table \ref{tab:sum_test_acc_cifar10}. We then prune these fully
trained networks.  Pruning decisions are made using a subset of the training
images, $I_{val}$.  Saliency according to the consituent metrics and myopic
oracle are computed using 256 and 512 random images from $I_{val}$ for CIFAR-10
and CIFAR-100 networks respectively.  The top-1 accuracy of the pruned networks
are measured using the entire test set, $I_{test}$.  The myopic oracle is
evaluated with $k=5,8,12,16$.

\section{Experimental Results and Discussion} \label{sec:results}

\begin{table*}[h]
\caption{Convolution weights (\%) removed for a 5\% accuracy drop on CIFAR-10 (gray) and CIFAR-100.}
\label{tab:sum_sparsity_5_cifar10}
\centering
\begin{tabular}{m{2.5cm}m{1cm}lllll}
\multicolumn{2}{c}{\textbf{Saliency Metric}} & \textbf{LeNet-5 } & \textbf{CIFAR10 } & \textbf{ResNet-20 } & \textbf{NIN } & \textbf{AlexNet } \\
\midrule

\multicolumn{2}{c}{\multirow{2}{3.5cm}{Mean of activations}}
& \cellcolor{gray!25}24$\pm$0.1  & \cellcolor{gray!25}29$\pm$2  &  \cellcolor{gray!25}6$\pm$1  & \cellcolor{gray!25}14$\pm$1  & \cellcolor{gray!25}37$\pm$6  \\
 &         & -                   & -                            & 2.6$\pm$0.5                  & 38.1$\pm$0.2                 & 22 $\pm$ 0.2               \\

\multicolumn{2}{c}{\multirow{2}{3.5cm}{1\textsuperscript{st} order Taylor expansion}}         
& \cellcolor{gray!25}33$\pm$6  & \cellcolor{gray!25}16$\pm$0.1  &  \cellcolor{gray!25}2$\pm0.1$  & \cellcolor{gray!25}14$\pm1$  & \cellcolor{gray!25}49$\pm6$  \\
 &         & -                   & -                            & 2.3$\pm$0.3                  & 38.2$\pm$0.1                 & 47 $\pm$ 0.4              \\

\multicolumn{2}{c}{\multirow{2}{3.5cm}{2\textsuperscript{nd} order Taylor expansion using Fisher information}}       
& \cellcolor{gray!25}22$\pm$5  & \cellcolor{gray!25}39$\pm$2    &  \cellcolor{gray!25}6$\pm3$  & \cellcolor{gray!25}21$\pm2$  & \cellcolor{gray!25}43$\pm6$  \\
 &         & -                   & -                            & 1.1$\pm$0.1                  & 38.5$\pm$0.4                 & 52 $\pm$ 6              \\

\multicolumn{2}{c}{\multirow{2}{3.5cm}{Average of gradients}}     
& \cellcolor{gray!25}25$\pm$6  & \cellcolor{gray!25}24$\pm$4    &  \cellcolor{gray!25}3$\pm0.3$  & \cellcolor{gray!25}24$\pm1$  & \cellcolor{gray!25}46$\pm9$  \\
 &         & -                   & -                            & 2.1$\pm$0.3                  & 36.7$\pm$0.3                 & 50 $\pm$ 7              \\

\multicolumn{2}{c}{\multirow{2}{3.5cm}{Mean squares of weights}}  
& \cellcolor{gray!25}17  & \cellcolor{gray!25}23                &  \cellcolor{gray!25}5  & \cellcolor{gray!25}28  & \cellcolor{gray!25}49  \\
 &         & -                   & -                            & 1.7                    & 40.7                   & 45               \\

\multirow{6}{2.5cm}{Myopic Oracle}  & \multirow{2}{1cm}{$k=5$}          
& \cellcolor{gray!25}29$\pm$5  & \cellcolor{gray!25}37$\pm$4    &  \cellcolor{gray!25}9$\pm4$  & \cellcolor{gray!25}31$\pm2$  & \cellcolor{gray!25}52$\pm0.3$  \\
 &                        & -             & -                   & 3.2$\pm$0.3                  & 39.5$\pm$0.4                 & 55 $\pm$ 7     \\

& \multirow{2}{1cm}{$k=8$}            
& \cellcolor{gray!25}26$\pm$3  & \cellcolor{gray!25}40$\pm$3    &  \cellcolor{gray!25}10$\pm4$  & \cellcolor{gray!25}31$\pm2$  & \cellcolor{gray!25}53$\pm5$  \\
 &                        & -             & -                   & 3.4$\pm$0.3                  & 39.8$\pm$0.4                 & 57 $\pm$ 7     \\

& \multirow{2}{1cm}{$k=12$}            
& \cellcolor{gray!25}27$\pm$4  & \cellcolor{gray!25}41$\pm$2    &  \cellcolor{gray!25}11$\pm4$  & \cellcolor{gray!25}32$\pm2$  & \cellcolor{gray!25}58$\pm3$  \\
 &                        & -             & -                   & 3.7$\pm$0.4                  & 39.9$\pm$0.4                 & 60 $\pm$ 0.8     \\

& \multirow{2}{1cm}{$k=16$}            
& \cellcolor{gray!25}26$\pm$7  & \cellcolor{gray!25}43$\pm$3    &  \cellcolor{gray!25}11$\pm3$  & \cellcolor{gray!25}33$\pm2$  & \cellcolor{gray!25}61$\pm4$  \\
 &                        & -             & -                   & 3.7$\pm$0.3                  & 40.0$\pm$0.6                 & 60 $\pm$ 1     \\

\end{tabular}

\end{table*}

Figure~\ref{fig:sensitivity_results} presents the result of our experimental
evaluation on the five chosen convolutional neural networks. For all five
networks, we see that the composite saliency metric matches or exceeds the
predictive quality of any of the individual constituent metrics until the test
accuracy of the network drops far below useful levels.

\subsection{Behaviour of Composite Metrics}
We would like to draw attention to the ResNet-20 (Figure~\ref{fig:resnet_20})
and NIN (Figure~\ref{fig:nin}) networks in particular.
For ResNet-20 (Figure~\ref{fig:resnet_20}), our experiment shows clearly that
some saliency metrics are very badly suited for guiding pruning on this
network. It is not that these are bad metrics; on the contrary, they perform
well on other networks.

However, the assumptions baked into these metrics are at odds with the reality
of the relationships of the weights, activations, and gradients in ResNet-20,
causing them to severely mispredict the effect on the loss function of pruning
any individual channel.Using the myopic oracle allows these metrics to be
excluded until their assumptions become more in line with the reality of the
network structure, instead of causing pathological behaviour if used
indiscriminately.

For NIN (Figure~\ref{fig:nin}), our experiment shows that the composition of
metrics via the oracle exhibits smooth, predictable behaviour, where the
individual metrics differ dramatically.

Even though the individual metrics have such
large differences, the composition of the metrics with the oracle is
well-behaved, leading to a much less damaging pruning that with any of the
metrics individually.

The remainder of the networks exhibit similar behaviour. For AlexNet
(Figure~\ref{fig:alexnet}), we see again that the composition of the saliency
metrics via the oracle yields a smooth, well-behaved metric, even though the
constituent metrics have large differences.

\subsection{Impact of k}

From Table \ref{tab:sum_sparsity_5_cifar10}, we can see that using a myopic
oracle can lead to a significant increase in the maximum number of weights but
only to a marginal increase when increasing k.  This trend would suggest that
the channel rankings given by the individual saliency metrics are often
accurate.  Considering more channels only offers a marginal improvement as the
least salient channels are often also ranked lowly by at least one of the
constitutent metrics, we only need to determine which saliency metric is
accurate for that pruning iteration.  Hence, choosing k to be equal to the
number of constitutent metrics allows us to choose between the saliency metrics
without inhibitively increasing cost of computation.

\subsection{Quality of Pruned Networks}

Table~\ref{tab:sum_sparsity_5_cifar10} summarizes the level of pruning achieved
in our experiments for a maximum reduction of 5\% points in top-1 test
accuracy.

Using the myopic oracle to compose existing saliency metrics yields a composite
metric which makes better pruning decisions than any of the individual metrics
which were composed. The myopic oracle consistently selects channels to prune
that result in a smaller loss in test accuracy. We also present the proportion
of weights removed for the constitutent saliency metrics, if used exclusively,
as in prior work.

On every network, our approach meets or exceeds the performance of all the
state of the art saliency metrics used individually. The best results are seen
on ResNet-20, where almost twice as many weights can be removed using our
approach versus the next-best individual saliency metric.

\section{Discussion}

\subsection{Pruning Scheme}

Combining saliency metrics with a myopic oracle within a simple pruning scheme
yields promising results.  However, more sophisticated saliency-based pruning
schemes can also take advantage of using multiple saliency metrics to remove
the maximum number of weights.  Our future work will cover testing the use of
multiple saliency metrics using a myopic oracle in other pruning schemes.

\subsection{Other Granularities}

The use of multiple saliency metrics is not limited to channel pruning, it can
apply to other granularities of pruning.  For example, with unstructured
pruning (fine-grain pruning) each saliency metric proposes a different set of
weights to be removed.  The myopic oracle can then choose which proposed set is
most accurate.
\section{Conclusion} \label{sec:conclusion}

Our method of composing multiple saliency metrics yields a composite metric that
significantly outperforms the individual constituent metrics. By dynamically
switching between different metrics based on the actual measured sensitivity of
the network, we avoid the occasional poor pruning decision made by even the
most advanced saliency metrics. Using our approach, data scientists are freed
from having to choose from a dizzying array of potential saliency metrics to
guide the pruning process.

By developing a method to dynamically switch between an arbitrary collection of
state-of-the-art saliency metrics based on their actual measured performance, we
can derive a composite metric with significantly improved performance, pruning
up to twice as many weights for the same drop in accuracy in our experiments.
Our approach advances the state of the art in identifying unnecessary or
redundant sets of neural network parameters.

\section*{Acknowledgement}
This work was supported by Science Foundation Ireland grant 13/RC/2094 to
Lero - The Irish Software Research Centre.  This work was also partly 
supported by Arm Research.
\bibliography{paper}

\end{document}